\documentclass[a4paper, 10pt, conference]{ieeeconf}      

\IEEEoverridecommandlockouts                              
\usepackage[table,xcdraw]{xcolor}
\usepackage{graphicx}
\usepackage{amsmath}
\usepackage{float}
\usepackage{cite}
\usepackage{tikz}

\newcommand\copyrighttext{%
	\footnotesize \copyright 2020 IEEE. Personal use of this material is permitted. Permission from IEEE must be obtained for all other uses, in any current or future media, including reprinting/republishing this material for advertising or promotional purposes, creating new collective works, for resale or redistribution to servers or lists, or reuse of any copyrighted component of this work in other works.}
\newcommand\copyrightnotice{%
	\begin{tikzpicture}[remember picture,overlay]
	\node[anchor=south,yshift=8pt,xshift=8pt] at (current page.south) {\fbox{\parbox{\dimexpr\textwidth-\fboxsep-\fboxrule\relax}{\copyrighttext}}};
	\end{tikzpicture}%
}

\title{\LARGE \bf
	Fast Object Classification and Meaningful Data Representation of Segmented Lidar Instances
}

\author{Lukas Hahn$^{1,2}$, Frederik Hasecke$^{1,2}$ and Anton Kummert$^{1}$
\thanks{$^{1}$ University of Wuppertal, Germany - Faculty of Electrical Engineering\newline
        { \tt\small firstname.lastname@uni-wuppertal.de}}
\thanks{$^{2}$ Aptiv, Wuppertal, Germany\newline
    	{ \tt\small firstname.lastname@aptiv.com}\vspace{0.4em}}
\thanks{\footnotesize This work is a result of the research project @CITY – Automated Cars and Intelligent Traffic in the City. The project is supported by the Federal Ministry for Economic Affairs and Energy (BMWi), based on a decision taken by the German Bundestag. The author is solely responsible for the content of this publication.}}

\begin{document}
\maketitle
\thispagestyle{empty}
\pagestyle{empty}
\copyrightnotice

\begin{abstract}
	
Object detection algorithms for Lidar data have seen numerous publications in recent years, reporting good results on dataset benchmarks oriented towards automotive requirements. Nevertheless, many of these are not deployable to embedded vehicle systems, as they require immense computational power to be executed close to real time.\\
In this work, we propose a way to facilitate real-time Lidar object classification on CPU. We show how our approach uses segmented object instances to extract important features, enabling a computationally efficient batch-wise classification. For this, we introduce a data representation which translates three-dimensional information into small image patches, using decomposed normal vector images. We couple this with dedicated object statistics to handle edge cases. We apply our method on the tasks of object detection and semantic segmentation, as well as the relatively new challenge of panoptic segmentation. Through evaluation, we show, that our algorithm is capable of producing good results on public data, while running in real time on CPU without using specific optimisation.
	
\end{abstract}

\section{INTRODUCTION}
\label{chapter:introduction}
Lidar sensors are used in a large number of fields, providing a three-dimensional representation of the given environment. Object detection on Lidar data is widely considered to be a crucial aspect for perception in automotive active safety and  autonomous driving. Hence, there is a large number of works on object detection \cite{Qi.2017, Zhou.2018, yan2018second, LangPointPillars} and odometry \cite{moosmann2011velodyne, behley2018efficient, LiLonet} in the Lidar space, with many approaches reporting incremental improvements on benchmark datasets \cite{Geiger.2012, behley2019iccv} targeted towards automotive applications.\\
While many methods show inventive data processing combined with deep learning techniques to achieve very good performance on those tasks, the vast majority is heavily optimised to these benchmarks and does not reach real-time performance, even when executed on powerful GPUs. This makes their application in vehicles unfeasible today and the foreseeable future, since dedicating so much computational power towards one algorithm is impractical concerning cost, power requirements and heat dissipation.\\

\begin{figure}[H]
	\centering
	\includegraphics[width=0.48\textwidth]{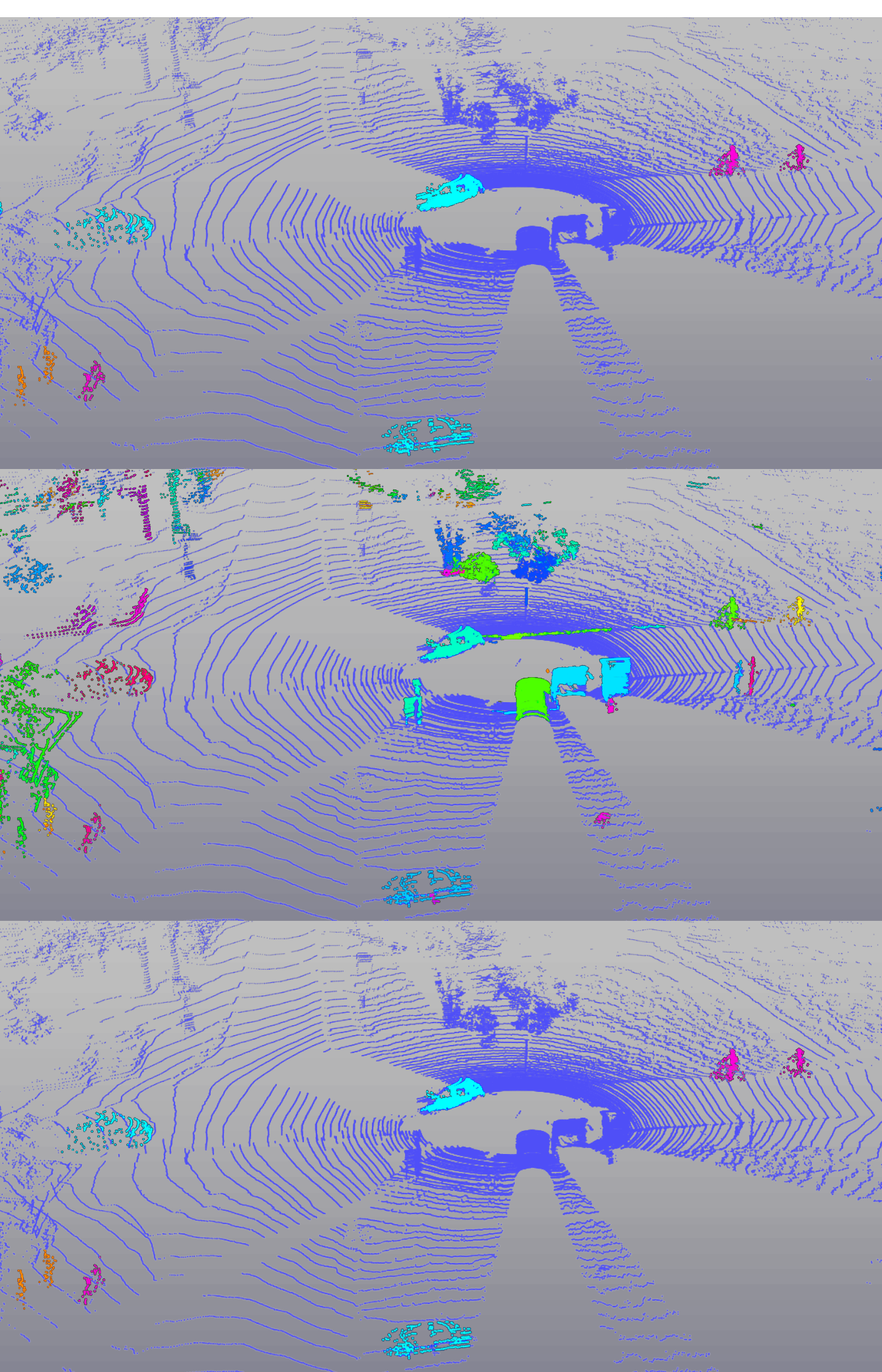}
	\caption{Exemplary street scene from the \textit{SemanticKITTI} dataset \cite{behley2019iccv}. Ground truth annotations \textit{(top)}, all segmented object instances after clustering with the algorithm of \cite{ourawesomeownwork} \textit{(middle)} and after classification with our approach \textit{(bottom)}. The proposed method is capable of effectively suppressing "None" object clusters and correctly classifying the relevant objects. Please note, that the clustering output is agnostic to object classes and the colours here are just used to differentiate instances.}
	\label{SemanticSample}
\end{figure}

We see promising results in recent development of methods for real-time object instance segmentation of Lidar point clouds \cite{ourawesomeownwork, Bogoslavskyi.2016}. Thus, in this work we propose a meaningful representation for object instances in Lidar sensor data, to facilitate a real-time capable, CPU-based classification.
We accomplish this task with a small custom neural network architecture, by applying methods to maintain three-dimensional image information in a two-dimensional representation and selective object statistics.

\section{RELATED WORK}
\label{chapter:relatedwork}
There is a multitude of publications concerned with object detection and classification on Lidar data. In an attempt to provide a short overview, we roughly divide them into three categories; algorithms that work on unregularised point clouds, those that use regularised ones, and algorithms that use fusion approaches combining Lidar with other sensors, mostly camera data.\\
A lot of methods process unregularised point clouds and usually generate per-point features, with PointNet \cite{Qi.2017}, offering more global feature generation, and PointNet++ \cite{Qi.2017b}, cascading instances of the aforementioned approach for more localised features, being the backbone for a larger number of publications including PointRCNN \cite{PointRCNN}. While facilitating good performance, those approaches are computationally intensive, which rules them out for embedded application.\\
A second group of algorithms regularises Lidar point clouds before processing them. To do so, grid structures are used in fixed \cite{Zhou.2018, shi2019pvrcnn, chen2019object} or variable sizes \cite{alsfasser}.  To alleviate the negative performance influence of many empty grid cells created by this methods, works like \cite{yan2018second} established specific network layers to exploit this sparsity,  providing a significant speed-up. \cite{LangPointPillars} reduce grids to a two-dimensional representation, enabling omission of costly 3D convolutions for faster runtime.\\
The third and last set of works adds information from camera sensors to enrich Lidar data. Some of them use camera images to create region proposals, which are refined by a Lidar algorithm in a subsequent step. Here, a popular approach is the use of a frustum for projection into the point cloud, as shown in \cite{FrustumPointNets, Zhao3D} and \cite{wang2019frustum} for example. Comparable methods utilise full three-dimensional object proposals from image inputs \cite{shin2019roarnet} or more complex deep learning fusion networks \cite{ChenMultiview}.\\
All of the approaches mentioned above are highly computationally intensive, with many of them not being real time capable even on high-end GPUs. There is not much work to be found concentrating on real-time application of Lidar algorithms on systems with less computational power. \cite{minemura2018lmnet} claim such capabilities using specific optimisations for powerful CPUs, but report sub-par results. 

\section{METHODS}
\label{chapter:methods}

To facilitate real-time CPU-based Lidar object classification of segmented instances, we consider the following aspects as valuable.

\subsection{Lidar data representation}
\label{DataRepresentation}
\begin{figure}[tb]
	\centering
	\includegraphics[width=0.36\textwidth]{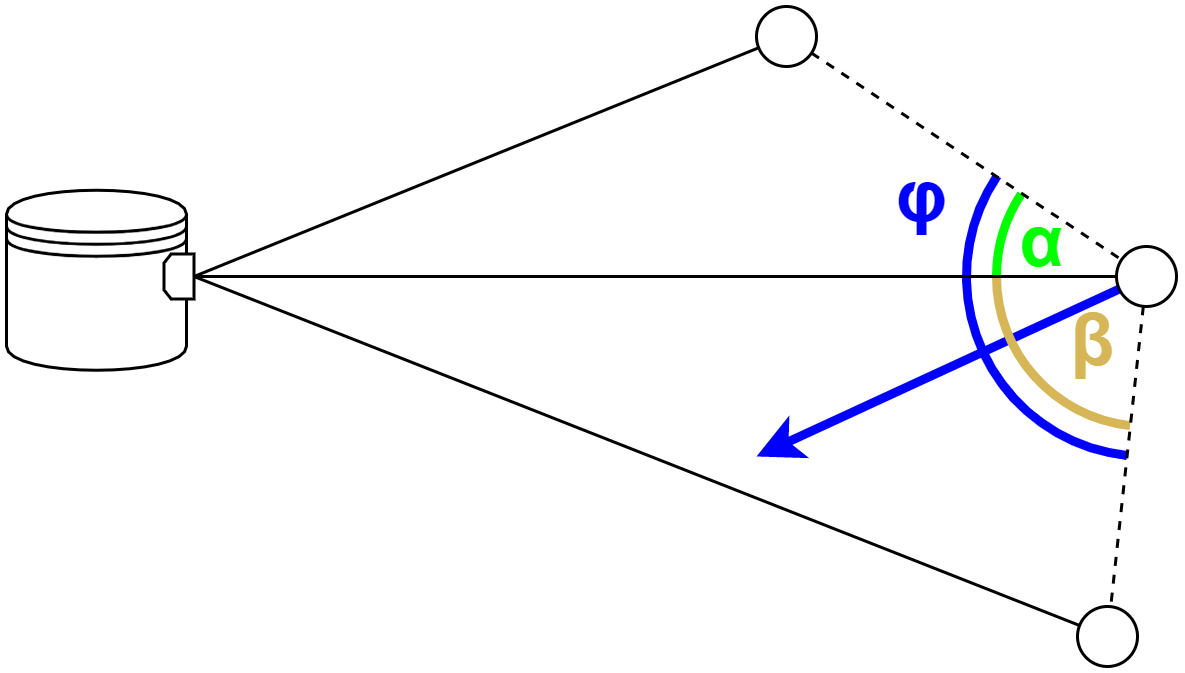}
	\caption{Relationship of angles between lines connecting adjacent Lidar measurement points and the line from the respective point to the sensor origin for calculating the a normal vector component. With $\phi = \alpha+\beta$, the angle bisector can be determined as $\frac{\phi}{2} = \frac{\alpha+\beta}{2}$.}
	\label{NormalVector}
\end{figure}

\begin{figure*}[tb]
	\centering
	\includegraphics[width=0.75\textwidth]{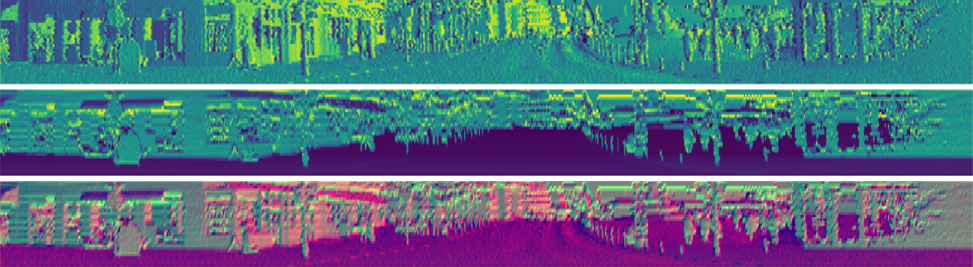}
	\caption{Visualisation of the horizontal \textit{(top)} and vertical \textit{(middle)} normal vector component image, as well as a combined view of both components \textit{(bottom)} for representative purpose.}
	\label{NormalVectorSamples}
\end{figure*}

From the raw data of a Lidar sensor, several different information aspects can be used to describe a measured point. The obvious first aspects are the X, Y and Z coordinate values of this point, usually given in Cartesian coordinates with the origin of the coordinate system in the location of the Lidar sensor. Derived from this, the distance or range of the point can be calculated, for example as Euclidean distance. Furthermore, the measured intensity of a point is normally given in the raw data, enabling indication of the surface reflectivity of an object.\\
To gain more knowledge about the surface of an object in the Lidar space, we propose to calculate an image representation of the horizontal and vertical component of the normal vector of each measured point. Such a representation is known in SLAM/odometry algorithms for Lidar data \cite{moosmann2011velodyne, behley2018efficient, LiLonet}. The normal vector for a measured Lidar point can be determined using the angle relationships shown in \textsc{Figure} \ref{NormalVector}. $\alpha$ and $\beta$ are the angles between the line from the sensor origin to the given point and the line from the given point to its respective neighbour. The angle bisector $\frac{\Phi}{2}$ of the combined angle $\phi = \alpha+\beta$ equals one component of the normal vector. We decided to use the scalar values of the angles instead of the more common cross product calculation between neighbouring points in the three-dimensional space. This strategy reduces the calculation of the normal vector image to a simple element-wise matrix subtraction.
To compute the horizontal component, the neighbouring points left and right of the point in question are used. For the vertical component, the neighbour above and below are considered respectively. Using just one neighbouring point here would allow for a potentially larger error. An exemplary representation of both components can be seen in \textsc{Figure} \ref{NormalVectorSamples}, with an additional combined view for illustration purposes.\\
To identify redundant information in these seven data representations and to reduce the number of input layers which later need to be processed, we conducted an ablation experiment testing all possible input combinations (see \textsc{Appendix}). It showed, that we are able to efficiently select only three layers, namely the intensity values and our horizontal and vertical normal vector component representations, and maintain performance with only a minimal accuracy decrease compared to using all seven of them.
\begin{figure*}[tb]
	\centering
	\includegraphics[width=0.75\textwidth]{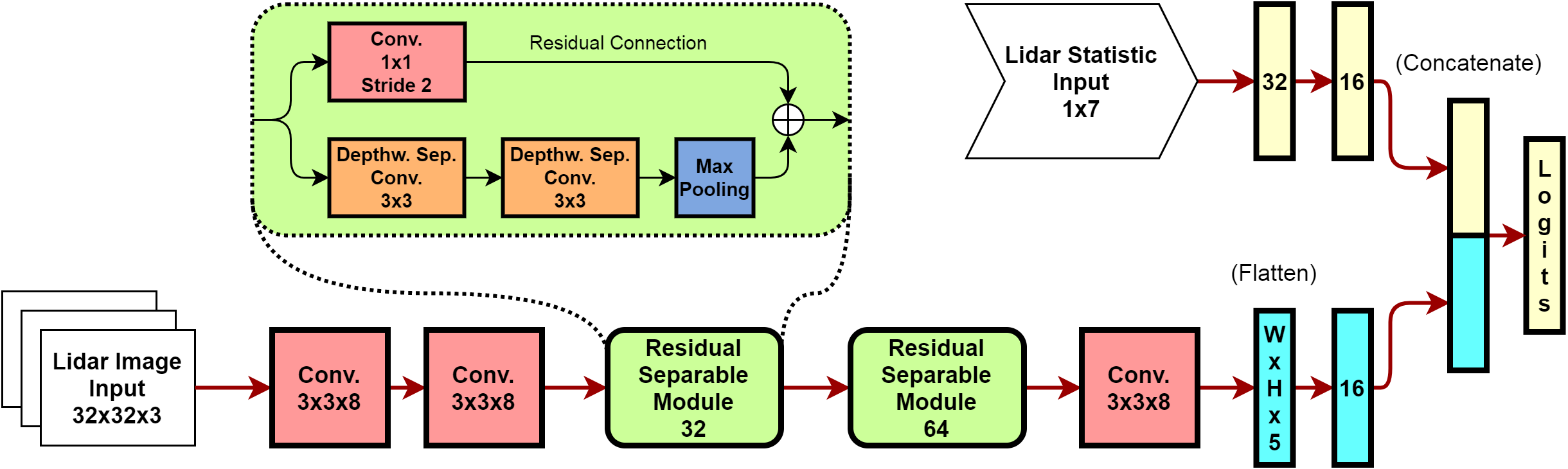}
	\caption{Architecture of the proposed CNN for fast Lidar object instance classification.}
	\label{LidarNetwork}
\end{figure*}

\subsection{Object instance and mask}
\label{InstanceAndVector}
\begin{figure}[tb]
	\centering
	\includegraphics[width=0.48\textwidth]{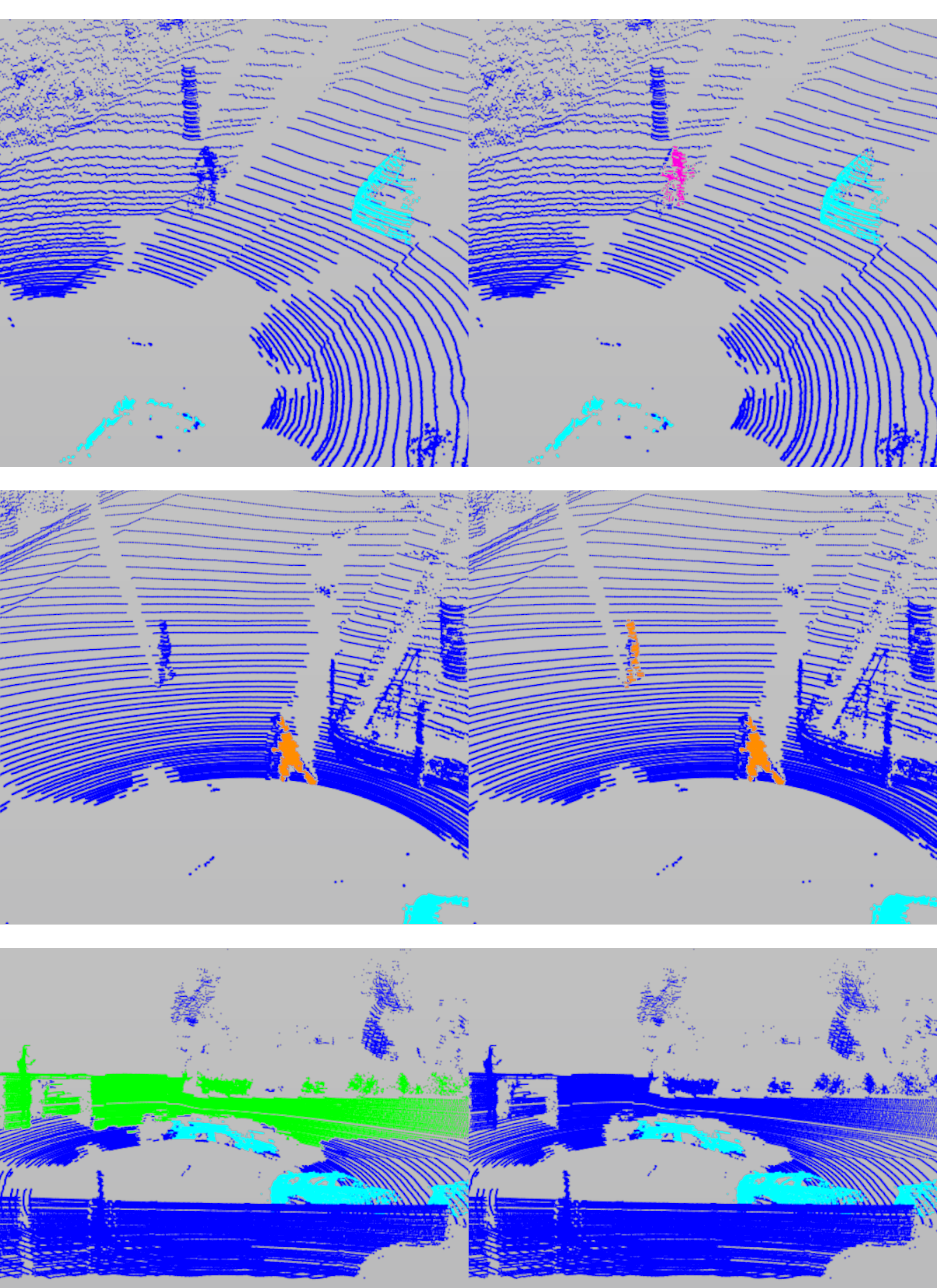}
	\caption{Influence of the additional statistic vector described in \ref{InstanceAndVector}. The \textit{top left} and \textit{middle left} figure show examples for false negatives and the \textit{bottom left} figure for false positives respectively. The additional statistic vector prevents some of these errors (\textit{right side}).}
	\label{StatisticInfluence}
\end{figure}

We build our classification approach on segmented object instances in the Lidar space. This facilitates a number of possible different applications. While integrating the proposed contributions of this work into a larger end-to-end neural network would be feasible, we considered the use of clustering algorithms to generate object instances. There are some methods available, which have been optimised to provide good performance in real-time Lidar instance segmentation on automotive data \cite{ourawesomeownwork, Bogoslavskyi.2016}. General purpose clustering algorithms like DBSCAN \cite{Ester.1996} could also be used, but are much more computationally intensive.\\
The use of the clustered objects for our classification is twofold; Primarily, we can apply the segmented instance as a mask on the different Lidar data representations described in \textsc{Section} \ref{DataRepresentation}, emphasising the points belonging to the object and eliminating background influence. Furthermore, we can extract statistical information about each segmented object without much difficulty. While there is no way to be completely certain, that an object is recognised in its full extent, such values are clearer and more deterministic than many machine learning representations and can be especially helpful in edge cases.\\
A clustering approach will segment all kind of object instances above a separated ground plane, independent of whether they belong to a supported class relevant for automotive application. Hence, our classification will be presented with a majority of "None" objects along the roadside. For example; walls of shops and houses can be confused with trucks, street lights or other poles can resemble pedestrians and in some cases, cars might even be confused with larger bushes.\\
We use the width, length and height of a segmented object, the number of Lidar points belonging to it, as well as the Euclidean and X- and Y-axis distance from the sensor origin. In this way we generate a vector of size seven. \cite{song2018classifying} use different but comparable geometric features in their approach to distinguish between main object categories in their closed-source dataset.\\
While influence on the overall classification performance scores is not particularly substantial, the use of our statistic vector proved to offer valuable decision support for critical edge cases in our experiments. Examples of this advantage in correcting both false positives and false negatives can be seen in \textsc{Figure} \ref{StatisticInfluence}.

\subsection{Object instance classification architecture}
Following our objective to facilitate a fast classification of Lidar instances, which is capable of running on CPU in real time, we propose a small Convolutional Neural Network (CNN) architecture. As depicted in \textsc{Figure} \ref{LidarNetwork}, it consists of two branches. The first one takes the statistic vector as input and comprises two fully connected layers. The second larger branch processes the Lidar image representations and is characterised by two residual separable modules in between common $3\times3$ convolution layers. Such a module features two depth-wise separable convolutions and maximum pooling in parallel to a residual connection with $1\times1$ filter kernel. Comparable structures have been popularised in deep learning architectures \cite{He.2016, Chollet.2017, Szegedy.2015} and shown benefits in application to small real-time networks \cite{Freeman.2018, howard2017mobilenets, Arriaga.2019}.\\
The complete CNN architecture has a total of $20\,802$ parameters, which results in a weight checkpoint file of $398$ kBytes in size. For comparison, state-of-the-art end-to-end Lidar networks have many millions of parameters and need much more memory to store weights accordingly, which is another cost driving factor in addition to their much higher computational requirements. 

\section{EXPERIMENTAL EVALUATION}
\label{chapter:evaluation}

To evaluate the performance of our proposed method, we consider different aspects and metrics. While good detection/classification rates are important, we developed our approach with a focus on real-time capability for CPU-based platforms. Hence, we are not aiming to set new benchmark high scores competing directly with magnitudes larger end-to-end Lidar networks. We rather aspire to provide reliable output given strong computational limitations.\\
We use the public dataset \textit{SemanticKITTI} by Behley et al. \cite{behley2019iccv}, which is an extension of the original \textit{KITTI} \cite{Geiger.2012} and provides semantic annotation for all sequences of its odometry benchmark, to make our evaluation transparent. An illustrative example of our method's performance can be seen in \textsc{Figure} \ref{SemanticSample}. The semantically segmented point clouds, as well as the additional instance labels in the dataset, allow for multiple pointwise evaluation approaches. Accordingly, we use three different tasks to assess the performance: 

\subsection{Semantic Segmentation}
\label{subsec:semantic}
For the semantic segmentation evaluation, we employ a general purpose clustering method to agnostically separate object instances in the three-dimensional point cloud. As described in \textsc{Section} \ref{chapter:methods}, we classify the segmented instances from these clusters.\\
Our approach classifies objects in five general automotive classes: ``Cars'', ``Trucks'', ``Pedestrians'', ``Bikes'' and the ``None'' class, which embodies all static background classes such as road surface, buildings and vegetation. To achieve this mapping, we combined the \textit{SemanticKITTI} classes ``Bicycle'', ``Bicyclist'', ``Motorcycle'' and ``Motorcyclist'' to ``Bike'', as well as ``Truck'', ``Bus'', ``On-Rails'' and ``Other-Vehicle'' to ``Trucks''. The classification network has been trained with the annotated point clouds of the available training logs. As suggested in the \textit{SemanticKITTI} API documentation, we kept the 8th log separate for validation. In this manner we are able to test semantic segmentation results with the reduced class mapping on unseen data. \textsc{Table} \ref{table:iou} shows the results for the class-wise semantic segmentation intersection over union (IoU) metric of the combined approach of clustering the point cloud and classifying each cluster separately. The IoU or \textit{Jaccard index} is defined as
\begin{equation*}
J(A, B) = \frac{\vert A \cap B \vert}{\vert A \cup B \vert} \:\: \widehat{=} \:\: \frac{TP}{TP+FP+FN}
\end{equation*}
and in practice can be described as the relation of true positives (TP) to the sum of TP, false positives (FP) and false negatives (FN).
This metric provides a good impression of pointwise segmentation quality, since the correct predictions and both types of incorrect predictions for each point are included in the equation.\\
The score of this metric is computed for two approaches. The first approach is the classification of clustered instances as an end-to-end pipeline on point cloud data to show the performance of the proposed method on unseen samples. The second approach applies the classification directly on the annotated ground truth instances in the dataset. This allows for a comparison, on how the performance is influenced by the grade of the provided object instances. As the results show, performance of the semantic segmentation directly depends on the quality of the given object clusters.\\

\begin{table}[tbp]
	\begin{center}
		\begin{tabular}[width=0.5\textwidth]{lccccc}
			\setlength{\tabcolsep}{10pt}
			\vspace{5pt}
			\textbf{Method}  & None & Car & Truck & Bike & Pedestrian\\ 
			\hline
			& & & & & \\
			Clustered Instances & 0.954 & 0.750 & 0.472 & 0.265 & 0.282\\ 
			& & & & & \\
			GT Instances & 0.994 & 0.926 & 0.732 & 0.525 & 0.558\\ 
			\hline
		\end{tabular} 
	\end{center}
	\caption{\textbf{Semantic Segmentation Results} as Intersection over Union (IoU) using clustered instances or ground truth instances as input for our classification approach.}
	\label{table:iou}
\end{table}

\vspace{-0.80em}
\subsection{Object Detection}
The second evaluation method is meant to assess the performance of object detection. For this, the previously mentioned IoU metric is used to define bins of precision in the detection. Ten bins are defined with a pointwise overlap of the ground truth objects and proposed clusters, ranging from an IoU of 0.5 to 0.95 in steps of 0.05. The average of all 10 bins is the single metric score, the Average Precision (AP), which is shown in \textsc{Table} \ref{table:ap}.\\ Additionally the AP for the overlap values of 0.5, 0.75 and 0.95 are listed, in which the evaluation is restricted to objects above the denoted IoU. 
\begin{table}[tbp]
	\begin{center}
		\begin{tabular}[width=0.5\textwidth]{lcccc}
			\setlength{\tabcolsep}{10pt}
			\vspace{5pt}
			\textbf{Method} & $AP$ & $AP^{0.5}$ & $AP^{0.75}$ & $AP^{0.95}$\\ 
			\hline
			& & & & \\
			Clustered Instances & 0.407 & 0.441 & 0.419 & 0.314 \\ 
			& & & & \\
			GT Instances & 0.554 & - & - & - \\ 
			\hline
		\end{tabular} 
	\end{center}
	\caption{\textbf{Object Detection Results} as Average Precision (AP) on clustered instances and provided ground truth instances.}
	\label{table:ap}
\vspace{-1.40em}
\end{table}
We adopt this metric definition from established benchmarks \cite{Lin2014MicrosoftCC, Cordts2016Cityscapes} for two- and three-dimensional bounding box object detection. 

\subsection{Panoptic Segmentation}
Given the nature of our method, class-less separation of object clusters and background followed by object classification, we can use it to perform the task of panoptic segmentation. This term was coined by Kirillov et al. in their work of the same name \cite{kirillov2019panoptic}. According to the authors, this task ``\textit{unifies the typically distinct tasks of semantic segmentation (assign a class label to each pixel) and instance segmentation (detect and segment each object instance)}''.\\
In panoptic segmentation the differentiation between \textit{Stuff} ie. background and \textit{Things}, in our case active road users, is as important as the separation of \textit{Things} among themselves. 
After separation into background and clusters and classification of the latter, we use the predicted class labels to remove the instance labels from clusters which are not part of the \textit{Things}, in our case all \textit{None} labels such as utility boxes, road signs and vegetation.\\
We validate our approach on the panoptic segmentation benchmark of the \textit{SemanticKITTI} dataset \cite{PanopticKITTI}. This challenge uses the panoptic quality (PQ), originally proposed by \cite{kirillov2019panoptic}, averaged over all classes $C$, as used by Porzi et al. \cite{porzi2019seamless}, on the whole test set. For a single class it is defined as
\begin{equation*}
PQ = \frac{\sum _{(\mathcal{S}, \hat{\mathcal{S}}) \in TP} IoU(\mathcal{S},\hat{\mathcal{S}}) }
{\vert TP \vert + \frac{1}{2} \vert FP \vert + \frac{1}{2} \vert FN \vert }
\end{equation*}
and is composed from the segmentation quality (SQ) and the recognition quality (RQ) as follows
\begin{equation*}
PQ = \underbrace{\frac{\sum _{(p, g) \in TP} IoU(p,g)}{\vert TP \vert}}_{\text{segmentation quality (SQ)}} \times \underbrace{\frac{\vert TP \vert}{\vert TP \vert + \frac{1}{2} \vert FP \vert + \frac{1}{2} \vert FN \vert}}_ {\text{recognition quality (RQ)}}.
\end{equation*}
At the time of writing, our method is the only submitted approach for the panoptic segmentation benchmark. We are positive, that at the time of publishing we will have dropped in this challenge, as our approach is as previously shown limited to only 4 of the 19 classes and will not yield competitive results summed over all classes. The results in \textsc{Table} \ref{table:pq} are therefore limited to the four classes described in \textsc{Section} \ref{chapter:methods}. We like to note, that our training set uses a reduced class mapping (see \textsc{Section} \ref{subsec:semantic}) and the reported official results are evaluated on the full class set. Therefore our performance on ``Truck'' and ``Bike'' suffers.

\begin{table}[tbp]
	\begin{center}
		\begin{tabular}[width=0.5\textwidth]{lccccc}
			\setlength{\tabcolsep}{10pt}
			\vspace{5pt}
			\textbf{Class} &\textbf{PQ}  &\textbf{SQ}  &\textbf{RQ}  &\textbf{IoU} \\ 
			\hline
			& & & & & \\
			Car & 0.754 & 0.866 & 0.87 & 0.792\\ 
			& & & & & \\
			Truck & 0.0534 & 0.888 & 0.0602 & 0.0371\\
			& & & & & \\
			Bike & 0.0822 & 0.723 & 0.114 & 0.0462\\ 
			& & & & & \\
			Pedestrian & 0.377 & 0.905 & 0.417 & 0.161\\ 
			& & & & & \\
			\hline
		\end{tabular} 
	\end{center}
	\caption{\textbf{Panoptic Segmentation Results} as panoptic quality (PQ), segmentation quality (SQ), recognition quality (RQ) and IoU using classification of clustered instances.}
	\label{table:pq}
\vspace{-2em}
\end{table}

\subsection{Timings}
We implemented the proposed network architecture in Python with Tensorflow, devoid of any further customisation or optimisation. For $100$ input instances of object proposals from the point cloud, which is representative of a residential area to inner-city scene, inference time on an Intel i7-6820HQ laptop CPU @ 2.70 GHz is $\approx32$ms. If execution is limited to only two threads, resembling a small embedded processor, this value rises to $\approx71$ms.\\
Timing measurements are often not published alongside high-performing publications. In exceptional cases, they are reported on powerful GPUs. To get a rough estimate about the runtime of state-of-the-art networks, we timed the inference of one point cloud from \textit{SemanticKITTI} with a network closely comparable to \textit{PointPillars} \cite{LangPointPillars}. Doing so took $\approx1.86$s, which is nearly $60$ times slower than our approach, on the same CPU.

\section{CONCLUSION}
\label{chapter:conclusion}

We have presented an algorithmic approach to facilitate real-time CPU-based classification of segmented object instances in Lidar data. Our approach uses a component-wise decomposed normal vector image, instance masks and selected statistics to create a meaningful data representation. The proposed CNN architecture is efficient in processing large batches of those instances, in a frequency high enough to enable real-time application while running on CPU, even in Python without specific optimisation. Through evaluation on public data we have shown, that our method achieves good performance on automotive Lidar semantic segmentation and object detection tasks, while being orders of magnitude faster to compute than current state-of-the-art approaches.\\
Through the combined use of instance segmentation and classification of these separated objects, we can further provide panoptic segmentation of street scenes. There is little previous work on that and to our best knowledge, ours is the first to report a method which can accomplish this task in real time on CPU. In future work, we aim to improve our detection accuracy by taking past time steps into account for refinement through causal tracking approaches.



\section*{APPENDIX}

\begin{table}[H]
	\begin{center}
		\begin{tabular}{ccccccc|c}
		\multicolumn{7}{c|}{Channel Configuration} &           \\
		X  & Y  & Z  & I  & D  & HNV & VNV & Test Acc. \\ \hline\hline \rowcolor[HTML]{86cfea}
		&    &    &    &    &     &     & 0.835     \\
		$\bullet$  & $\bullet$  & $\bullet$  &    &    &     &     & 0.875     \\ \rowcolor[HTML]{86cfea}
		$\bullet$  & $\bullet$  & $\bullet$  & $\bullet$  &    &     &     & 0.892     \\ 
		$\bullet$  & $\bullet$  & $\bullet$  &    & $\bullet$  &     &     & 0.879     \\ \rowcolor[HTML]{86cfea}
		$\bullet$  & $\bullet$  & $\bullet$  & $\bullet$  & $\bullet$  &     &     & 0.895     \\ 
		$\bullet$  & $\bullet$  & $\bullet$  &    &    & $\bullet$   & $\bullet$   & 0.882     \\ \rowcolor[HTML]{86cfea}
		$\bullet$  & $\bullet$  & $\bullet$  & $\bullet$  &    & $\bullet$   & $\bullet$   & \textbf{0.900}     \\ 
		$\bullet$  & $\bullet$  & $\bullet$  &    & $\bullet$  & $\bullet$   & $\bullet$   & 0.890     \\ \rowcolor[HTML]{86cfea}
		$\bullet$  & $\bullet$  & $\bullet$  & $\bullet$  & $\bullet$  & $\bullet$   & $\bullet$   & \textbf{0.902}     \\ 
		&    &    & $\bullet$  &    &     &     & 0.887     \\ \rowcolor[HTML]{86cfea}
		&    &    &    & $\bullet$  &     &     & 0.861     \\ 
		&    &    & $\bullet$  & $\bullet$  &     &     & 0.891     \\ \rowcolor[HTML]{86cfea}
		&    &    & $\bullet$  &    & $\bullet$   & $\bullet$   & \textbf{0.896}     \\ 
		&    &    &    & $\bullet$  & $\bullet$   & $\bullet$   & 0.881     \\ \rowcolor[HTML]{86cfea}
		&    &    & $\bullet$  & $\bullet$  & $\bullet$   & $\bullet$   & 0.894     \\ 
		&    &    &    &    & $\bullet$   & $\bullet$   & 0.873     
		\end{tabular}
	\end{center}
	\caption{\textbf{Results of the input channel ablation experiment.} Configuration options include the Cartesian coordinates (X, Y, Z), intensity (I) and depth (D), as well as the horizontal and vertical normal vector component images (HNV, VNV). A binary mask of reflected lidar points is applied at all times.}
\end{table}


\bibliography{root}
\bibliographystyle{ieeetr}

\end{document}